\documentclass[letterpaper]{article}
\usepackage{aaai2026}  
\usepackage{times}                
\usepackage{helvet}               
\usepackage{courier}              
\usepackage[hyphens]{url}
\usepackage{graphicx}             
\usepackage{natbib}               
\usepackage{caption}              
\usepackage{algorithm}
\usepackage{algorithmic}
\usepackage{amsmath}        
\usepackage{amssymb}        
\usepackage{multirow}
\usepackage{booktabs}
\usepackage[hyphens]{url}
\title{ReviBranch: Deep Reinforcement Learning for Branch-and-Bound 

with Revived Trajectories}

\author{
  Jiabao Dou\textsuperscript{$\dagger$}, 
  Jiayi Nie\textsuperscript{$\dagger$}, 
  Yihang Cheng\textsuperscript{$\ddagger$}, 
  Jinwei Liu\textsuperscript{$\dagger$}, \\
  Yingrui Ji\textsuperscript{$|$}, 
  Canran Xiao\textsuperscript{$\P$}, 
  Feixiang Du\textsuperscript{$\S$},
  Jiaping Xiao\textsuperscript{*}
}

\affiliations{
  \textsuperscript{$\dagger$}Hong Kong Baptist University\\
  \textsuperscript{$\ddagger$}Central China Normal University\\
  \textsuperscript{$\P$}Central South University\\
  \textsuperscript{$|$}University of Chinese Academy of Sciences\\
  \textsuperscript{$\S$}Tongling University\\
  \textsuperscript{*}Nanyang Technological University\\[0.5em]
}

\begin{document}
\maketitle

\begin{abstract}
The Branch-and-bound (B\&B) algorithm is the main solver for Mixed Integer Linear Programs (MILPs), where the selection of branching variable is essential to computational efficiency. However, traditional heuristics for branching often fail to generalize across heterogeneous problem instances, while existing learning-based methods such as imitation learning (IL) suffers from dependence on expert demonstration quality, and reinforcement learning (RL) struggles with limitations in sparse rewards and dynamic state representation challenges. To address these issues, we propose ReviBranch, a novel deep RL framework that constructs revived trajectories by reviving explicit historical correspondences between branching decisions and their corresponding graph states along search-tree paths. During training, ReviBranch enables agents to learn from complete structural evolution and temporal dependencies within the branching process. Additionally, we introduce an importance-weighted reward redistribution mechanism that transforms sparse terminal rewards into dense stepwise feedback, addressing the sparse reward challenge. Extensive experiments on different MILP benchmarks demonstrate that ReviBranch outperforms state-of-the-art RL methods, reducing B\&B nodes by $4.0\%$ and LP iterations by $2.2\%$ on large-scale instances. The results highlight the robustness and generalizability of ReviBranch across heterogeneous MILP problem classes.
\end{abstract}

\section{Introduction}

Mixed Integer Linear Programming (MILP) provides a fundamental framework for modeling combinatorial optimization problems across production planning, scheduling, facility location, and network design \cite{wolsey1999integer}. The NP-hard nature of these problems requires sophisticated solution methods, with the branch-and-bound (B\&B) algorithm serving as the core algorithmic framework in modern MILP solvers \cite{kianfar2010branch}.

The B\&B algorithm employs a systematic divide-and-conquer strategy, recursively partitioning the solution space through tree search \cite{land2009automatic}. At each node of this search tree, a critical decision must be made: select which fractional integer variable to branch \cite{wolsey1999integer}. Crucially, the choice of branching strategy profoundly impacts the algorithm's performance \cite{achterberg2008constraint}. Different strategies can lead to vastly different search tree sizes, computational complexity, and ultimately, solution quality and solving time \cite{patel2007active, morrison2016branch}. Consequently, developing effective branching strategies is essential for solving complex real-world problems. However, traditional branching heuristics often prove inefficient due to their reliance on simple local criteria without considering global optimization context \cite{achterberg2005branching}.

\begin{figure*}
  \centering
  \includegraphics[width=1\linewidth]{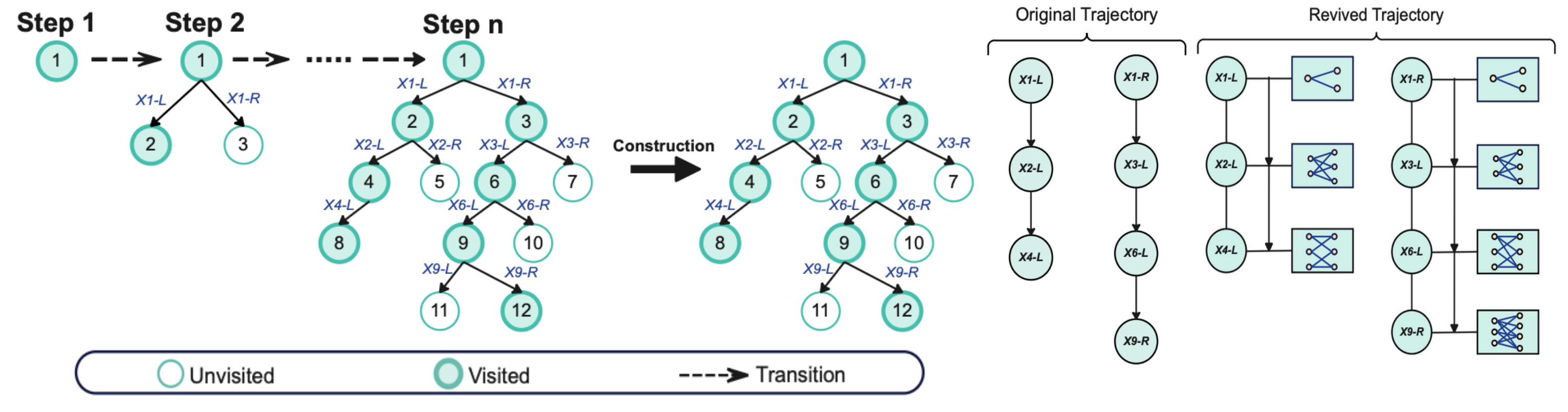}
  \caption{Branch-and-Bound Algorithm and Trajectory Construction. Left: B\&B search tree showing branching decisions and state transitions, where each node displays the selected branching variable with constraints $x_i \leq$ (left) and $x_i \geq$ (right). The action represents the choice of branching variable $x_i$. Right: Comparison between Original Trajectory (action sequences only) and Revived Trajectory (complete state-action correspondences with graph structures).}
  \label{fig:bnb_trajectory}
\end{figure*}

Recent advances in machine learning have introduced sophisticated approaches to branching. Imitation learning (IL) methods train neural networks to mimic expert decisions \cite{khalil2016learning, gasse2019exact}, but rely on expensive expert demonstrations \cite{he2014learning}. Reinforcement learning (RL) approaches overcome this through autonomous exploration \cite{etheve2020reinforcement, parsonson2023reinforcement}, but face three fundamental challenges: \textit{(1) Long-term dependency modeling}: Early branching choices may only reveal their quality after hundreds of subsequent decisions, creating significant credit assignment difficulties. \textit{(2) Dynamic state representation challenges}: Each branching decision fundamentally alters the entire search tree structure, with variables getting fixed, constraints becoming active/inactive, and graph topology changing dynamically \cite{khalil2017learning}. This creates difficulties in defining stable state representations that can capture evolving structural information throughout the branching process, as traditional methods struggle to model these complex state transitions effectively \cite{bengio2021machine}. \textit{(3) Sparse reward signals}: RL agents typically receive rewards only upon complete problem resolution, as the first integer feasible solution is often found very deep in the search tree. This sparse feedback severely hinders learning efficiency and credit assignment during training \cite{parsonson2023reinforcement, sutton1998reinforcement}.

To address these limitations, we consider leveraging historical trajectory information to enable agents to learn from past branching decisions. However, during reinforcement learning training, agents inevitably generate suboptimal trajectories through $\epsilon$-greedy exploration strategies \cite{sutton1998reinforcement}. Relying solely on historical action sequences may cause agents to learn erroneous patterns by mistaking local optima for global ones \cite{schaul2015prioritized}. Therefore, we propose ReviBranch, a deep reinforcement learning framework that addresses these challenges through novel architectural and algorithmic innovations.

Specifically, our ReviBranch has three key innovations:
\textbf{(1)} It employs an Encoder-Revival-Decoder architecture that combines BipartiteGCN \cite{gasse2019exact} for structural graph representation with Graph-Sequence Decoder featuring multidirectional cross-attention mechanisms, enabling agents to capture both spatial problem structure and temporal decision dependencies, addressing Challenge 1.
\textbf{(2)} It constructs revived trajectories by reviving historical graph states from stored data through distributed storage and reconstructing complete graph state-action correspondences (Figure~\ref{fig:bnb_trajectory}), resolving dynamic state representation challenges and enabling agents to learn from precise temporal context, addressing Challenge 2.
\textbf{(3)} It develops Importance-Weighted Reward Redistribution (IWRR) that transforms sparse terminal rewards into dense stepwise feedback through temporal importance weighting, enabling effective credit assignment for individual branching decisions and addressing the sparse reward problem, addressing Challenge 3.

\section{Related Work}

\paragraph{Classical Branching Heuristics} Early approaches include Most Fractional Branching, which selects variables closest to 0.5, and Random Branching for baseline comparison. Pseudocost branching (PB) \cite{benichou1971experiments} uses historical pseudocosts, enabling rapid decisions but suffering from cold starts and poor cross-instance adaptability. Strong branching (SB) \cite{linderoth1999computational} overcomes the myopia of PB via costly one-step lookahead, achieving superior solution quality, but at computational costs often orders of magnitude higher than PB.

\paragraph{Learning to Branch} To bypass SB's prohibitive online cost, IL methods learn from SB demonstrations. Khalil et al. \cite{khalil2016learning} pioneered this direction using support vector machines to replicate SB rankings, but required costly per-instance training. Gasse et al. \cite{gasse2019exact} introduced bipartite Graph Convolutional Networks that capture variable-constraint relationships through message passing, learning SB policies offline. Zarpellon et al. \cite{zarpellon2021parameterizing} explicitly parameterized search tree states to improve cross-instance generalization. However, these imitation learning approaches remain fundamentally constrained by their expert teachers' performance ceiling. RL offers potential for discovering superior branching strategies through autonomous exploration. Etheve et al. \cite{etheve2020reinforcement} proposed Fixed-depth Monte Carlo Tree Search (FMSTS) to reduce episode length through depth-first trajectory construction. Parsonson et al. \cite{parsonson2023reinforcement} developed retro branching by decomposing search trees into subtree trajectories, reducing partial observability but maintaining sparse reward challenges. However, current approaches lack explicit temporal dependency modeling for sequential branching decisions and fail to establish accurate mappings between decisions and evolving states.

\section{Background}
\paragraph{Mixed Integer Linear Programming} MILP is an optimization problem in which values are assigned to decision variables, subject to linear constraints, so as to minimize a linear objective function. MILPs can be formulated in the following standard form \cite{wolsey1999integer}:
\begin{equation}
\min_{x} c^T x \quad \text{s.t.} \quad Ax \leq b, \quad l \leq x \leq u, \quad x_j \in \mathbb{Z} \text{ for } j \in \mathcal{I}
\end{equation}
where $c \in \mathbb{R}^{n_0}$ is the objective coefficient vector, $A \in \mathbb{R}^{m \times n_0}$ is the constraint coefficient matrix, $b \in \mathbb{R}^m$ is the constraint right-hand side vector, $l, u \in \mathbb{R}^{n_0}$ are the lower and upper variable bound vectors, and $\mathcal{I} \subseteq \{1,2,\ldots,n_0\}$ denotes the set of indices for integer-constrained variables. To solve MILPs exactly, the B\&B algorithm is widely used \cite{land2009automatic}.

\paragraph{Branch-and-Bound Algorithm} B\&B algorithm systematically explores the solution space through recursive partitioning \cite{land2009automatic}. At each node, the algorithm solves the LP relaxation, selects a fractional integer variable $x_j$ from $\mathcal{F}(x^*) = \{i \in \mathcal{I} : x_i^* \notin \mathbb{Z}\}$, and creates two child nodes with constraints $x_j \leq \lfloor x_j^* \rfloor$ and $x_j \geq \lceil x_j^* \rceil$, where $x_j^*$ is the fractional value in the current LP solution. SB \cite{alvarez2017machine} evaluates each candidate by solving LP relaxations for both potential child nodes, while PB \cite{benichou1971experiments} uses historical statistics to estimate branching effects efficiently, avoiding SB's computational overhead. This motivates machine learning approaches for intelligent variable selection.

\paragraph{Graph Representation for MILP} Following Gasse et al. \cite{gasse2019exact}, MILP instances are represented as bipartite graphs with constraint and variable vertices, connected when $A_{ij} \neq 0$. Constraint nodes encode right-hand side values and slack information, while variable nodes capture bounds, objective coefficients, and LP solution values. BipartiteGCN processes such graphs through message passing to learn structural representations.

\paragraph{RL Framework}Sequential decision-making problems can be formulated as Markov Decision Processes (MDPs) defined by the tuple $(\mathcal{S}, \mathcal{A}, \mathcal{P}, \mathcal{R}, \gamma)$, where $\mathcal{S}$ represents the state space, $\mathcal{A}$ the action space, $\mathcal{P}$ the transition probabilities, $\mathcal{R}$ the reward function, and $\gamma$ the discount factor. Q-learning \cite{watkins1992q} aims to learn the optimal action-value function $Q^*(s,a)$ that represents the expected cumulative reward for taking action $a$ in state $s$ and following the optimal policy thereafter. Deep Q-Networks (DQN) \cite{mnih2015human} address this limitation by approximating the optimal action-value function using neural networks $Q(s,a;\theta) \approx Q^*(s,a)$, where $\theta$ represents the network parameters. DQN utilizes experience replay and target networks for training stability.

\section{Method}
We present ReviBranch, a deep reinforcement learning framework that employs an Encoder-Revival-Decoder neural network architecture trained via DQN. Our framework introduces three key innovations: (1) Revived Trajectories Construction, which revives historical graph states to construct enhanced trajectories with complete state-action correspondences; (2) Graph-Sequence Decoder, which combines structural graph features with temporal decision sequences through multi-directional attention mechanisms; and (3) IWRR, which transforms sparse terminal rewards into dense stepwise feedback through temporal importance weighting.

\subsection{Problem Formulation} 
We formulate the branching variable selection problem as an episodic MDP, where each episode corresponds to solving one MILP instance through a sequence of branching decisions until optimality is reached.

\textbf{States.} At each branching step $i$, the state $s_i$ is represented as a bipartite graph $G = (C \cup V, E)$, where $C$ and $V$ represent constraint and variable nodes with 5- and 19-dimensional features respectively, and $E$ captures relationships when $A_{ij} \neq 0$. The complete features are provided in Appendix B. The state includes the current action set $\mathcal{A}(s_i)$ of fractional integer variables eligible for branching.

\textbf{Actions.} The action space $\mathcal{A}(s_i) = \{j \in \mathcal{I} : x_j^* \notin \mathbb{Z}\}$ consists of fractional integer variables eligible for branching, where $x_j^*$ is the LP relaxation solution value. Our DQN agent employs a policy $\pi(a_i|s_i)$ to select a variable index $a_i \in \mathcal{A}(s_i)$ for branching.

\textbf{Transitions.} The transition function $\mathcal{P}(s_{i+1}|s_i, a_i)$ is deterministic and follows the standard branch-and-bound mechanism. When variable $j$ is selected for branching at state $s_i$, the environment creates two child nodes with additional constraints $x_j \leq \lfloor x_j^* \rfloor$ and $x_j \geq \lceil x_j^* \rceil$, where $x_j^*$ is the fractional LP solution value.

\begin{figure*}
  \centering
  \includegraphics[width=1\linewidth]{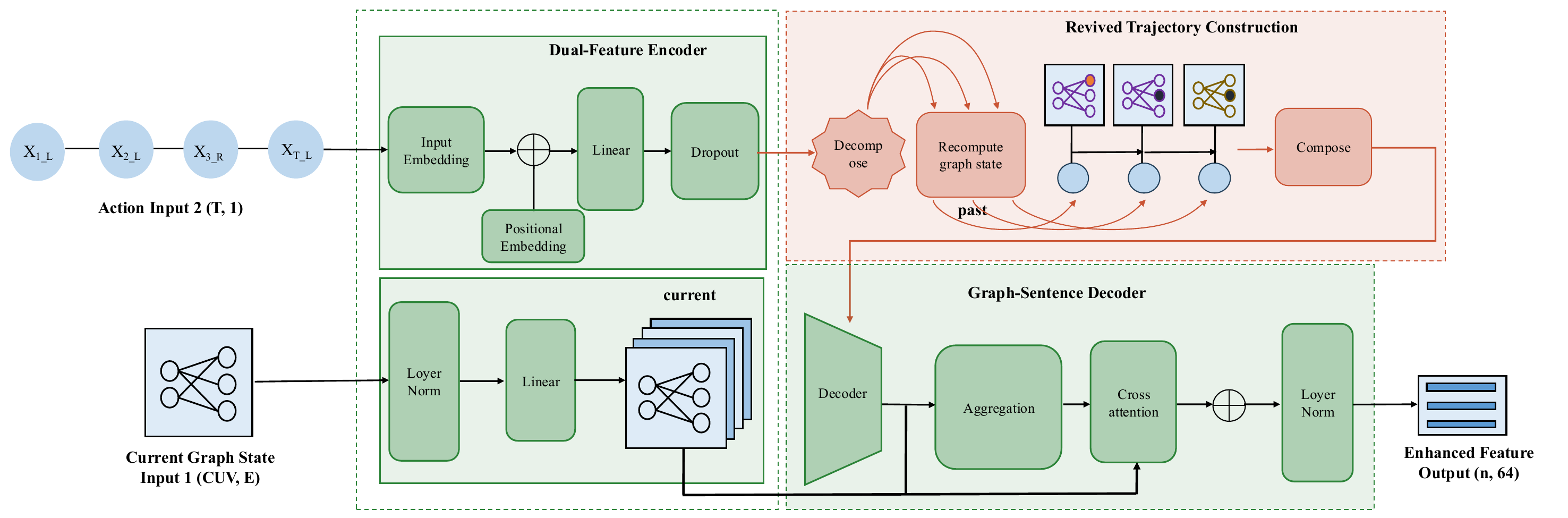}
  \caption{Architecture of ReviBranch framework.}
  \label{fig:revibranch_architecture}
\end{figure*}

\textbf{Revived Trajectories.} As illustrated in Figure~\ref{fig:bnb_trajectory}, we construct revived trajectories $\tau^{\text{revived}} = \{(s_0^{\text{hist}}, a_0), (s_1^{\text{hist}}, a_1), \ldots, (s_{L-1}^{\text{hist}}, a_{L-1})\}$ (where $L$ denotes the trajectory length) by reviving historical graph states $s_i^{\text{hist}}$ from stored data and pairing them with corresponding actions $a_i$ (where $a_0$ denotes the first branching decision, $a_1$ the second, and so forth). This enables the reconstruction of complete state-action correspondences that capture both structural evolution and temporal dependencies within the search tree.

\textbf{Reward.} We introduce an importance-weighted reward redistribution mechanism. The base reward assigns $r = -1$ at each branching step and $r = 0$ when the action leads to subtree pruning, which is then redistributed using temporal importance weights to provide dense learning signals throughout the episode.

\subsection{ReviBranch Architecture}
Our ReviBranch framework employs a novel Encoder-Revival-Decoder architecture that processes current MILP state and historical context through three main components (Figure~\ref{fig:revibranch_architecture}):

\paragraph{Dual-Feature Encoder}
The encoder consists of two parallel streams processing different input modalities. The first stream employs BipartiteGCN to extract structural features from the bipartite graph, transforming the discrete optimization structure into continuous variable embeddings $\mathbf{V} \in \mathbb{R}^{n_i \times d}$ that serve as memory context for the Transformer decoder. The second stream processes historical action sequences $\mathbf{a}_{0:L-1} = [a_0, a_1, \ldots, a_{L-1}]$ of variable lengths through a learned embedding mechanism. Each action index $a_i$ representing a selected branching variable is mapped to a $d$-dimensional embedding vector through learned action embeddings. Positional encodings are added to preserve temporal order information in the sequence representations. The encoded sequence undergoes layer normalization to produce preliminary trajectory representations $\mathbf{R}_{action} \in \mathbb{R}^{L \times d}$ (action representation matrix) containing action embeddings with positional information, which will be used in the revival process to construct complete step-level representations.

\paragraph{Revived Trajectories Construction}
In DQN training, the agent stores historical experiences in a replay buffer, where each transition contains $(s_i, a_i, r_i, s_{i+1})$. A general approach would store complete historical context within each transition, requiring each of the $L$ transitions to store up to $L$ historical states, leading to $\mathcal{O}(L^2)$ storage complexity for a trajectory of length $L$, which is prohibitive for large-scale MILPs. Our revived trajectories construction reduces storage complexity from $\mathcal{O}(L^2)$ to $\mathcal{O}(L)$ through distributed storage. Each transition stores its own complete graph state data $G_i = (C_i, V_i, E_i)$ and action $a_i$, enabling distributed reconstruction of complete trajectories during training.

\begin{algorithm}[tb]
\caption{Revived Trajectories Construction}
\label{alg:revival}
\textbf{Input}: Sampled trajectory $\tau{=}(s_0,a_0,s_1,a_1,\ldots,s_{L{-}1})$ from replay buffer\\
\textbf{Output}: Complete revived trajectory

$\mathbf{R}_{\text{traj}}{=}[\mathbf{R}_0,\mathbf{R}_1,\ldots,\mathbf{R}_{L{-}1}]{\in}\mathbb{R}^{L{\times}d}$
\begin{algorithmic}[1]
    \FOR{$i{=}0$ \TO $L{-}1$}
        \STATE $G_i{\leftarrow}\text{RetrieveStoredState}(s_i)$ \hfill \COMMENT{Get stored graph}
        \STATE $\mathbf{V}_i{\leftarrow}\text{BipartiteGCN}(G_i)$ \hfill \COMMENT{Reconstruct graph embeddings}
        \STATE $\mathbf{a}_i^{\text{emb}}{\leftarrow}\text{ActionEmb}(a_i){+}\text{PE}(i)$ \hfill \COMMENT{Generate action embeddings}

        \STATE \textbf{// Step-Level Representation Construction}
        \STATE $\mathbf{v}_{\text{mean}}{\leftarrow}\frac{1}{n_i}\sum_{j{=}1}^{n_i}\mathbf{V}_i[j]$ \hfill \COMMENT{Mean pooling}
        \STATE $\mathbf{v}_{\text{max}}{\leftarrow}\max_{j{=}1}^{n_i}\mathbf{V}_i[j]$ \hfill \COMMENT{Max pooling}
        \STATE $\mathbf{w}{\leftarrow}\text{softmax}(\text{MLP}(\mathbf{V}_i))$ \hfill \COMMENT{Attention weights}
        \STATE $\mathbf{v}_{\text{att}}{\leftarrow}\sum_{j{=}1}^{n_i}\mathbf{w}[j]{\cdot}\mathbf{V}_i[j]$ \hfill \COMMENT{Attention pooling}
        \STATE $\mathbf{v}_{\text{combined}}{\leftarrow}\text{MLP}([\mathbf{v}_{\text{mean}};\mathbf{v}_{\text{max}};\mathbf{v}_{\text{att}}])$ \hfill \COMMENT{Fuse pooled features}
        \STATE $\mathbf{R}_i{\leftarrow}\text{MLP}([\mathbf{v}_{\text{combined}};\mathbf{a}_i^{\text{emb}}])$ \hfill \COMMENT{Fuse with action}
    \ENDFOR
    \STATE \textbf{return} $\mathbf{R}_{\text{traj}}{=}[\mathbf{R}_0,\mathbf{R}_1,\ldots,\mathbf{R}_{L{-}1}]$
\end{algorithmic}
\end{algorithm}

During training, given a sampled trajectory $\tau = (s_0, a_0, s_1, a_1, \ldots, s_{L-1})$ from the replay buffer, the revival module reconstructs complete graph state-action correspondences. For each historical step $i$, the module retrieves the stored graph state data $G_i$ and reconstructs the original graph embeddings $\mathbf{V}_i = \text{BipartiteGCN}(G_i)$. Simultaneously, action embeddings are generated: $\mathbf{a}_i^{emb} = \text{ActionEmb}(a_i) + \text{PE}(i)$, where ActionEmb() is a learned embedding lookup and PE() denotes positional encoding. Step-level representations are then constructed by fusing graph and action information through StepBuilder, which aggregates variable embeddings from the graph state and combines them with the corresponding action embedding to create a unified step representation (detailed fusion mechanism in Algorithm 1):
\begin{equation}
\mathbf{R}_i = \text{StepBuilder}(\mathbf{V}_i, \mathbf{a}_i^{emb})
\end{equation}

This revival process is essential because MILP graph structures evolve dynamically during branch-and-bound traversal, where variables get fixed, constraints become active/inactive, and the graph topology changes with each branching decision. The completed revived trajectories $\tau^{\text{revived}} = \{(s_0^{\text{hist}}, a_0), (s_1^{\text{hist}}, a_1), \ldots, (s_{L-1}^{\text{hist}}, a_{L-1})\}$ capture complete graph state-action correspondences while preserving the full structural evolution of the optimization process.

\paragraph{Graph-Sequence Decoder}
The decoder employs a Transformer decoder architecture with revived trajectory sequence $\mathbf{R}_{traj} \in \mathbb{R}^{L \times d}$ as target input and variable embeddings $\mathbf{V} \in \mathbb{R}^{n_i \times d}$ as memory context. The decoder processes the revived trajectory sequence and performs cross-attention with the variable memory to capture trajectory-variable interactions.

The decoder output undergoes aggregation to extract a unified trajectory representation $\mathbf{R}_{unified} \in \mathbb{R}^{d}$ through weighted averaging of the sequence. Multi-directional cross-attention consolidates information through three complementary pathways:
\begin{align}
\mathbf{E}_1 &= \mathrm{Attn}(\mathbf{R}_{unified}, \mathbf{V}, \mathbf{V}) \in \mathbb{R}^{d} \quad \text{(traj-to-var)} \\
\mathbf{E}_2 &= \mathrm{Attn}(\mathbf{V}, \mathbf{R}_{unified}, \mathbf{R}_{unified}) \in \mathbb{R}^{n_i \times d} \quad \text{(var-to-traj)} \\
\mathbf{E}_3 &= \mathrm{Attn}(\mathbf{V}, \mathbf{R}_{traj}, \mathbf{R}_{traj}) \in \mathbb{R}^{n_i \times d} \quad \text{(var-to-seq)}
\end{align}
The first pathway enables trajectory-to-variable attention for global context understanding. The second pathway allows variable-to-trajectory attention for trajectory-aware variable characterization. The third pathway captures fine-grained variable-to-sequence interactions for temporal pattern modeling. Enhanced variable embeddings are computed through multi-directional cross-attention fusion:
\begin{align}
\mathbf{V}_{enhanced} &= \mathbf{V} + \alpha \cdot \text{Fusion}(\mathbf{E}_1, \mathbf{E}_2, \mathbf{E}_3)
\end{align}
where $\alpha$ is a learned scaling parameter. The fusion function is defined as $\text{Fusion}(\mathbf{E}_1, \mathbf{E}_2, \mathbf{E}_3) = \text{MLP}([\mathbf{E}_1; \mathbf{E}_2; \mathbf{E}_3])$, which concatenates and transforms the multi-directional attention outputs, producing enhanced variable embeddings $\mathbf{V}_{enhanced} \in \mathbb{R}^{n_i \times d}$ for Q-value computation.

\paragraph{IWRR} We address sparse rewards by redistributing terminal rewards across all steps.At episode termination, we compute temporal importance weights $w_i = (L-i)/L$ to prioritize early decisions, then redistribute terminal rewards as:
\begin{equation}
r_i^{\text{dense}} = R_{\text{terminal}} \times w_i
\end{equation}
where $R_{\text{terminal}} = \sum_{j=0}^{L-1} r_j$ is the cumulative episode reward (sum of base step rewards), $L$ is the episode length, and $i \in \{0, 1, \ldots, L-1\}$ is the step index. The redistributed rewards are normalized and mapped to $[-0.9, -0.1]$:
\begin{equation}
r_i^{\text{final}} = -0.9 + 0.8 \times \frac{w_i - \frac{1}{L}}{1 - \frac{1}{L}}
\end{equation}
This temporal weighting mechanism provides dense learning signals while maintaining training stability.

\paragraph{Training Procedure}
Our ReviBranch framework employs standard DQN training with experience replay. During each episode, the agent collects transitions $(s_i, a_i, r_i, s_{i+1})$ while maintaining action histories and storing graph state data. At episode termination, sparse environmental rewards are transformed into dense signals through Importance-Weighted Reward Redistribution and stored in the replay buffer. For network training, mini-batches are sampled from the buffer, revived trajectories are constructed via the revival module, and the network is updated using standard DQN loss with dense rewards:

\begin{equation}
\mathcal{L} = \mathbb{E}[(r_i^{\text{dense}} + \gamma \max_{a'} Q_{\text{target}}(s_{i+1}, a') - Q(s_i, a_i))^2]
\end{equation}
Complete training algorithm is provided in Appendix A.

\section{Experiments \& Results}
We conduct comparative experiments evaluating ReviBranch against the IL approach of Gasse et al. (2019), RL methods (the state-of-the-art Retro Branching), and traditional heuristic baselines. We also conduct ablation studies to validate each component. The source code is included in the supplementary material and will be released on GitHub upon paper acceptance.

\subsection{Experimental Setup}

\textbf{Environment Configuration.} We employ SCIP 7.0.0 \cite{Achterberg2009} as the backend mixed-integer programming solver, interfaced through the Ecole framework \cite{prouvost2020ecole}. We set a time limit of one hour per instance, disable solver restarts, and restrict cutting plane generation to the root node only. The complete environment configuration is provided in Appendix A.

\textbf{Benchmark.} We evaluate on three NP-hard combinatorial optimization benchmarks with three difficulty levels each: \textbf{Set Covering} instances: Easy (200$\times$400), Medium (500$\times$1000), Hard (1000$\times$1500). \textbf{Combinatorial Auction} instances: Easy (20$\times$60), Medium (30$\times$80), Hard (50$\times$100). \textbf{Capacitated Facility Location} instances: Easy (10$\times$15), Medium (25$\times$25), Hard (40$\times$30).

\textbf{Training.} We train ReviBranch using Deep Q-Network with prioritized experience replay on each benchmark separately, using the medium difficulty level for training. Training instances are dynamically generated during reinforcement learning using Ecole's instance generators with random seeds. Testing uses pre-generated instances with distinct seeds to ensure no data leakage and proper generalization assessment. Training is set to a maximum of 600,000 epochs with early stopping based on validation performance. Detailed training configurations, hyperparameters, hardware specifications and training curve are provided in Appendix A.

\begin{table*}[t]
\centering
\begin{tabular}{l|cc|cc|cc}
\toprule
\multicolumn{7}{c}{\textbf{Set Covering}} \\
\midrule
\multirow{2}{*}{Method} & \multicolumn{2}{c|}{Easy} & \multicolumn{2}{c|}{Medium} & \multicolumn{2}{c}{Hard} \\
& LP$\downarrow$ & Nodes$\downarrow$ & LP$\downarrow$ & Nodes$\downarrow$ & LP$\downarrow$ & Nodes$\downarrow$ \\
\midrule
SB & 847.3±15.2\% & 19.6±17.9\% & 5835±16.9\% & 119.1±19.4\% & 17259.5±18.1\% & 532.4±19.9\% \\
SCIP Default & 1046.9±8.7\% & 24.8±8.1\% & 7051±11.2\% & 186.4±13.2\% & 21675.6±14.4\% & 746.4±16.3\% \\
\midrule
PB & 1494.6±20.8\% & 43.1±26.0\% & 11807±18.2\% & 315.3±22.8\% & 28406.8±21.8\% & 889.1±23.1\% \\
IL & 973.5±13.9\% & 23.2±16.4\% & 6334±15.1\% & 161±20.9\% & 18796.1±18.6\% & 638.6±19.4\% \\
Retro & \textbf{895.7±15.7\%}$\downarrow$ & \textbf{22.4±13.1\%}$\downarrow$ & 6307.4±19.6\% & 159.6±21.8\% & 19102.5±16.2\% & 657.2±23.1\% \\
\textbf{Revi} & 962.9±17.3\% & 25.9±19.8\% & \textbf{6095±12.7\%}$\downarrow$ & \textbf{151.7±17.3\%}$\downarrow$ & \textbf{18031.7±19.4\%}$\downarrow$ & \textbf{583.7±20.8\%}$\downarrow$ \\
\midrule
\multicolumn{7}{c}{\textbf{Capacitated Facility Location}} \\
\midrule
\multirow{2}{*}{Method} & \multicolumn{2}{c|}{Easy} & \multicolumn{2}{c|}{Medium} & \multicolumn{2}{c}{Hard} \\
& LP$\downarrow$ & Nodes$\downarrow$ & LP$\downarrow$ & Nodes$\downarrow$ & LP$\downarrow$ & Nodes$\downarrow$ \\
\midrule
SB & 46.4±11.7\% & 16.6±12.1\% & 168.1±14.1\% & 62.5±13.9\% & 515.2±13.1\% & 153.8±10.7\% \\
SCIP Default & 53.6±13.6\% & 19.1±11.3\% & 185.7±10.0\% & 64.3±11.8\% & 624.9±9.6\% & 184.5±11.0\% \\
\midrule
PB & 50.5±13.6\% & 17.9±16.1\% & 177.4±12.7\% & 67.2±15.8\% & 577.9±15.9\% & 180.1±16.8\% \\
IL & 45.7±9.3\% & 16.8±9.9\% & \textbf{159.8±10.5\%}$\downarrow$ & \textbf{57.7±12.4\%}$\downarrow$ & 542.6±12.4\% & 165.9±10.5\% \\
Retro & \textbf{46.3±12.9\%}$\downarrow$ & \textbf{16.2±11.0\%}$\downarrow$ & 162.2±9.3\% & 58.6±10.1\% & 533.8±11.7\% & 161.2±10.9\% \\
\textbf{Revi} & 49.6±10.3\% & 17.3±12.7\% & 171.7±12.7\% & 61.2±11.3\% & \textbf{529.7±12.6\%}$\downarrow$ & \textbf{159.9±11.4\%}$\downarrow$ \\
\midrule
\multicolumn{7}{c}{\textbf{Combinatorial Auction}} \\
\midrule
\multirow{2}{*}{Method} & \multicolumn{2}{c|}{Easy} & \multicolumn{2}{c|}{Medium} & \multicolumn{2}{c}{Hard} \\
& LP$\downarrow$ & Nodes$\downarrow$ & LP$\downarrow$ & Nodes$\downarrow$ & LP$\downarrow$ & Nodes$\downarrow$ \\
\midrule
SB & 34.7±13.1\% & 11.9±10.3\% & 115.8±15.6\% & 32.0±12.1\% & 526.8±18.9\% & 197.3±15.7\% \\
SCIP Default & 47.8±15.3\% & 18.7±12.9\% & 136.5±19.4\% & 45.1±15.2\% & 637.9±15.5\% & 231.0±17.3\% \\
\midrule
PB & 74.1±21.4\% & 28.1±23.7\% & 226.5±23.2\% & 88.9±25.3\% & 1210.6±20.8\% & 562.3±24.9\% \\
IL & \textbf{40.9±16.1\%}$\downarrow$ & \textbf{12.1±13.5\%}$\downarrow$ & 131.9±17.6\% & 36.8±16.3\% & 569.8±18.2\% & 205.3±16.9\% \\
Retro & 41.6±18.3\% & 12.5±15.7\% & 126.4±15.0\% & 36.3±14.7\% & 552.1±13.6\% &  203.6±13.8\% \\
\textbf{Revi} & 45.3± 17.4\% & 13.8±14.2\% & \textbf{125.9± 19.7\%}$\downarrow$ & \textbf{35.7±17.6\%}$\downarrow$ & \textbf{547.8± 16.5\%}$\downarrow$ & \textbf{200.9±15.7\%}$\downarrow$ \\
\bottomrule
\end{tabular}
\caption{Performance comparison of branching methods on three MILP benchmarks across difficulty levels. Results show geometric mean ± standard deviation of LP iterations and B\&B nodes over 50 test instances per difficulty level. \textbf{Lower values indicate better performance}. Bold indicates best performance among learning-based methods.}
\label{tab:benchmark_results}
\end{table*}

\textbf{Testing.} Each method is evaluated on every testing instance using 5 different random seeds. For each benchmark, 50 testing instances per difficulty level (easy/medium/hard) are randomly generated with fixed seeds, totaling 150 instances per problem class, which were unseen during training. We report the geometric mean of LP iterations and branch-and-bound nodes as primary metrics, following standard practice in MILP literature. LP iterations serve as a proxy for computational efficiency, as fewer LP iterations typically correlate with reduced inference time per branching decision. All metrics include average per-instance standard deviation (in percentage).

\textbf{Baseline Methods.} We compare ReviBranch against representative branching strategies spanning traditional heuristics and machine learning approaches. \textbf{Machine learning baselines:} Imitation learning (IL) \cite{gasse2019exact} and Retro Branching \cite{parsonson2023reinforcement}. \textbf{Traditional baselines:} Strong Branching (SB) \cite{alvarez2017machine}, Pseudocost Branching (PB) \cite{benichou1971experiments}, and SCIP's default reliability pseudocost branching.

\subsection{Results \& Analysis}

\paragraph{Performance Comparison with RL Baselines}
Table~\ref{tab:benchmark_results} reveals several critical insights about ReviBranch's performance characteristics across different problem and difficulty levels. A key finding is that ReviBranch exhibits a distinctive performance pattern that varies with problem complexity, fundamentally differing from the retro branching method. On Set Covering problems, the performance gap between ReviBranch and Retro increases substantially as complexity grows: from ReviBranch performing marginally worse on Easy instances (962.9 vs 895.7 LP iterations, representing a 7.5\% increase, worse performance) to achieving substantial advantages on Hard instances (18031.7 vs 19102.5, achieving a 5.6\% improvement, better performance). This progressive performance improvement is even more pronounced in the number of tree nodes, where ReviBranch achieves 11.2\% fewer nodes on Hard instances (583.7 vs 657.2). These results indicate that simple instances can be effectively solved through the local optimization strategies employed by retro branching, whereas complex instances require the global temporal reasoning capabilities that ReviBranch provides.

This pattern is validated by trajectory length sensitivity analysis, which shows that longer trajectories improve performance. The model with trajectory length 25 generates fewer nodes and shorter inference time than trajectory length 10, and as trajectory length gradually increases, both nodes and inference time further decrease.

Therefore, ReviBranch's benefits become apparent when branching decisions exhibit strong sequential dependencies, where early branching choices significantly influence the effectiveness of later decisions through constraint propagation and search space reduction. This characteristic intensifies with problem complexity as larger instances create deeper interdependencies between variables. In contrast, simpler instances often allow for more independent branching decisions, where each choice has limited impact on subsequent decisions. In such cases, ReviBranch's temporal modeling becomes less advantageous compared to direct heuristics focusing on immediate local variable properties rather than historical patterns.

Overall, across medium and hard difficulty levels (large-scale instances) on all three benchmarks, ReviBranch achieves an average reduction of 4.0\% in branch-and-bound nodes and 2.2\% in LP iterations compared to state-of-the-art reinforcement learning methods.

\paragraph{Comparison with other methods} ReviBranch consistently outperforms both traditional heuristics and learning-based methods across most benchmarks. While SB achieves the best performance with 4.5\% fewer LP iterations than ReviBranch on Hard Set Covering instances, ReviBranch significantly outperforms PB by 36.5\% and IL baselines by 4.1\% on the same instances, with similar advantages observed across other benchmarks. Notably, ReviBranch demonstrates superior performance over IL methods on large-scale instances, where the benefits of temporal reasoning become more pronounced.

\paragraph{Generalization analysis} Despite training only on medium-scale instances, ReviBranch demonstrates reasonable generalization capabilities across two dimensions.
Our method exhibits effective adaptability when problem complexity increases. For Set Covering, ReviBranch shows progressively expanding performance advantages from medium instances (LP: 6095 vs 6307 Retro) to hard instances (LP: 18031.7 vs 19102.5 Retro). For Capacitated Facility Location, it maintains competitive performance across difficulty levels, demonstrating robustness to problem scale variations. Additionally, ReviBranch achieves strong performance across three structurally distinct benchmarks, indicating that ReviBranch learns generalizable branching strategies that transfer across different optimization landscapes.

\begin{figure}[h]
\centering
\includegraphics[width=0.95\linewidth]{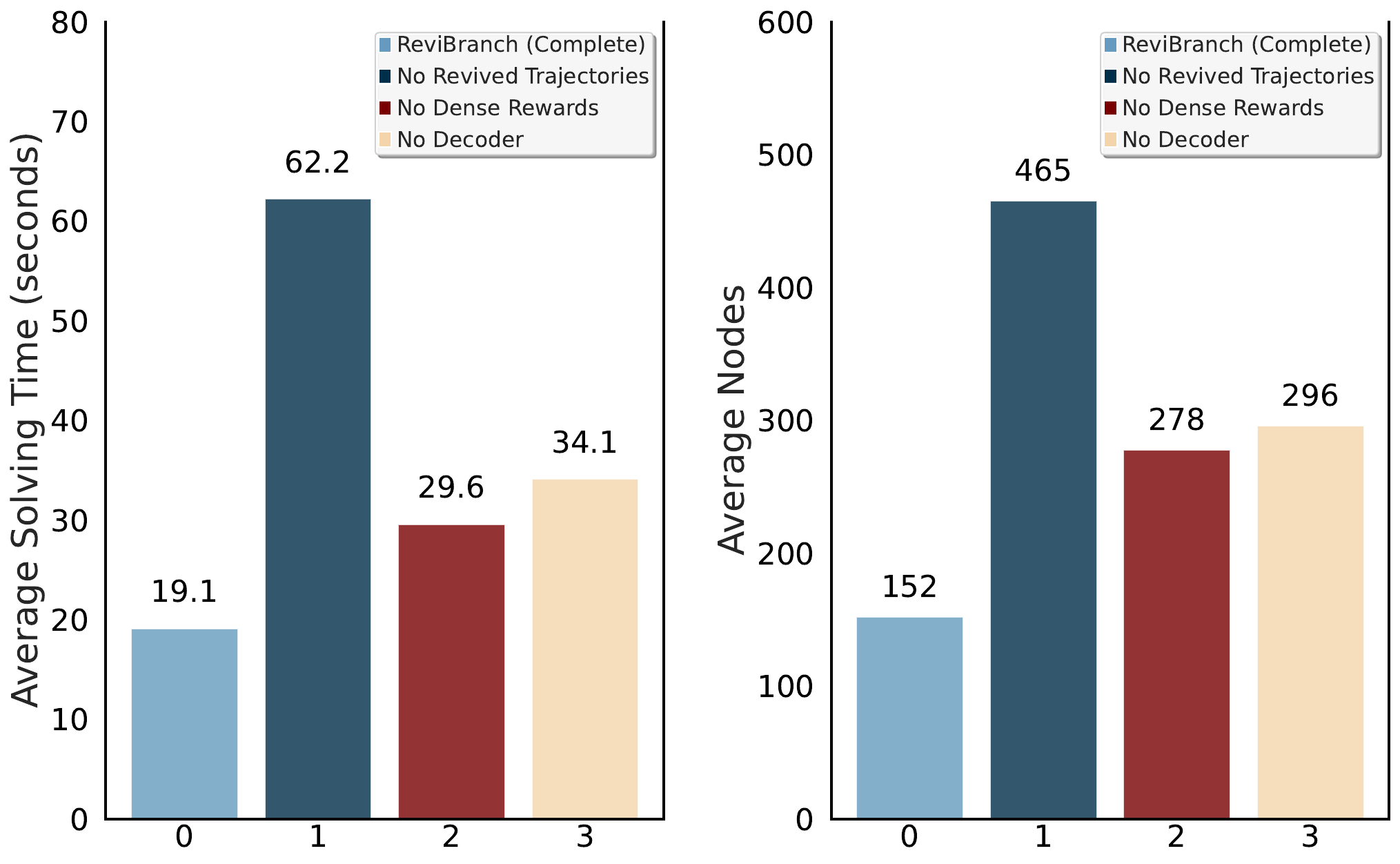}
\caption{Ablation study results on Set Covering instances. Performance comparison of three ablation variants: (1) No Revived Trajectories, (2) No Dense Rewards, and (3) No Decoder. Results show average solving time and nodes with standard deviation over 50 medium-scale instances across three random seeds.}
\label{fig:ablation}
\end{figure}

\paragraph{Ablation Experiments} We design three ablation variants: \textbf{no revived trajectories}, \textbf{no dense rewards}, and \textbf{no decoder}. To ensure statistical reliability of our results, we train each variant with three different random seeds and report average performance on 50 medium Set Covering instances. As shown in Figure~\ref{fig:ablation}, the no revived trajectories variant exhibits significant performance degradation: solving time increases by 110.1\% compared to no dense rewards and 82.5\% compared to no decoder; nodes increase by 67.3\% compared to no dense rewards and 57.1\% compared to no decoder. This occurs because removing revived trajectories prevents tracking graph evolution, leading to poor variable selection and prolonged solving time. Meanwhile, exploration stability deteriorates, with standard deviation increasing by 125\% (time) and 79\% (nodes) relative to no dense rewards, and 69\% (time) and 34\% (nodes) relative to no decoder. In contrast, the no dense rewards and no decoder variants show moderate performance decline, indicating that dense rewards and decoder components, while valuable, are non-decisive and primarily provide technical support for effective utilization of revived trajectories.

\begin{figure}[!t]
\centering
\includegraphics[width=0.95\linewidth]{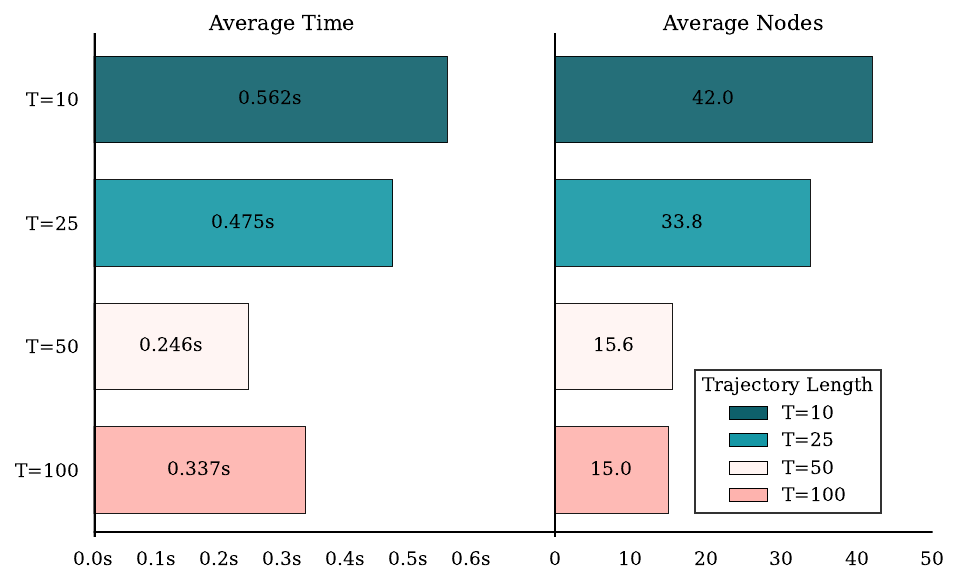}
\caption{Trajectory length sensitivity analysis on Combinatorial Auction instances (20×60 scale). Results show the trade-off between model performance (nodes) and computational efficiency (inference time) across different trajectory lengths (T=10, 25, 50, 100).}
\label{fig:trajectory_analysis}
\end{figure}

\paragraph{Trajectory length sensitivity analysis}
We conduct trajectory length sensitivity analysis to understand its impact on ReviBranch performance. As shown in Figure~\ref{fig:trajectory_analysis}, we test four different trajectory length configurations (T=10, 25, 50, 100) on 20×60 scale Combinatorial Auction instances. The experimental results show that the model with trajectory length 10 performs worst, with an average inference time of 0.562 seconds and requiring 42.0 nodes. Model performance further improves as trajectory length increases from 25 to 100.

However, we find that although the trajectory length 100 model has 0.6 fewer nodes than trajectory length 50, the inference time of trajectory length 100 is 0.357 seconds, which is 45\% longer than that of trajectory length 50 (0.246 seconds). As the problem complexity increases, this performance pattern requires increasing computational costs. This overhead affects inference time, as each action selection requires recomputing the entire historical trajectory.

\section{Conclusion}
We propose ReviBranch, a deep reinforcement learning framework that advances MILP branching through revived trajectory construction. By establishing explicit mappings between branching decisions and evolving graph states, ReviBranch addresses fundamental challenges: sparse rewards, long-term dependencies, and dynamic state representation. Our experiments demonstrate that ReviBranch outperforms state-of-the-art RL methods on large-scale instances, reducing branch-and-bound nodes by 4.0\% and LP iterations by 2.2\%. The framework exhibits strong generalization across diverse problem classes, with advantages most pronounced on complex instances.

ReviBranch establishes temporal reasoning as an effective paradigm for sequential optimization decisions. Future work may explore adaptive trajectory selection mechanisms and cross-trajectory reward attribution, while extending the approach to other solver components and sequential decision problems in combinatorial optimization.

\bibliography{references}

\makeatletter
\@ifundefined{isChecklistMainFile}{
  \newif\ifreproStandalone
  \reproStandalonetrue
}{
  \newif\ifreproStandalone
  \reproStandalonefalse
}
\makeatother

\ifreproStandalone
\documentclass[letterpaper]{article}
\usepackage[submission]{aaai2026}
\setlength{\pdfpagewidth}{8.5in}
\setlength{\pdfpageheight}{11in}
\usepackage{times}
\usepackage{helvet}
\usepackage{courier}
\usepackage{xcolor}
\frenchspacing

\begin{document}
\fi
\setlength{\leftmargini}{20pt}
\makeatletter\def\@listi{\leftmargin\leftmargini \topsep .5em \parsep .5em \itemsep .5em}
\def\@listii{\leftmargin\leftmarginii \labelwidth\leftmarginii \advance\labelwidth-\labelsep \topsep .4em \parsep .4em \itemsep .4em}
\def\@listiii{\leftmargin\leftmarginiii \labelwidth\leftmarginiii \advance\labelwidth-\labelsep \topsep .4em \parsep .4em \itemsep .4em}\makeatother

\setcounter{secnumdepth}{0}
\renewcommand\thesubsection{\arabic{subsection}}
\renewcommand\labelenumi{\thesubsection.\arabic{enumi}}

\newcounter{checksubsection}
\newcounter{checkitem}[checksubsection]

\newcommand{\checksubsection}[1]{%
  \refstepcounter{checksubsection}%
  \paragraph{\arabic{checksubsection}. #1}%
  \setcounter{checkitem}{0}%
}

\newcommand{\checkitem}{%
  \refstepcounter{checkitem}%
  \item[\arabic{checksubsection}.\arabic{checkitem}.]%
}
\newcommand{\question}[2]{\normalcolor\checkitem #1 #2 \color{blue}}
\newcommand{\ifyespoints}[1]{\makebox[0pt][l]{\hspace{-15pt}\normalcolor #1}}

\section*{Reproducibility Checklist}

\vspace{1em}
\hrule
\vspace{1em}

\textbf{Instructions for Authors:}

This document outlines key aspects for assessing reproducibility. Please provide your input by editing this \texttt{.tex} file directly.

For each question (that applies), replace the ``Type your response here'' text with your answer.

\vspace{1em}
\noindent
\textbf{Example:} If a question appears as
\begin{center}
\noindent
\begin{minipage}{.9\linewidth}
\ttfamily\raggedright
\string\question \{Proofs of all novel claims are included\} \{(yes/partial/no)\} \\
Type your response here
\end{minipage}
\end{center}
you would change it to:
\begin{center}
\noindent
\begin{minipage}{.9\linewidth}
\ttfamily\raggedright
\string\question \{Proofs of all novel claims are included\} \{(yes/partial/no)\} \\
yes
\end{minipage}
\end{center}
Please make sure to:
\begin{itemize}\setlength{\itemsep}{.1em}
\item Replace ONLY the ``Type your response here'' text and nothing else.
\item Use one of the options listed for that question (e.g., \textbf{yes}, \textbf{no}, \textbf{partial}, or \textbf{NA}).
\item \textbf{Not} modify any other part of the \texttt{\string\question} command or any other lines in this document.\\
\end{itemize}

You can \texttt{\string\input} this .tex file right before \texttt{\string\end\{document\}} of your main file or compile it as a stand-alone document. Check the instructions on your conference's website to see if you will be asked to provide this checklist with your paper or separately.

\vspace{1em}
\hrule
\vspace{1em}


\checksubsection{General Paper Structure}
\begin{itemize}

\question{Includes a conceptual outline and/or pseudocode description of AI methods introduced}{(yes/partial/no/NA)}
yes

\question{Clearly delineates statements that are opinions, hypothesis, and speculation from objective facts and results}{(yes/no)}
yes

\question{Provides well-marked pedagogical references for less-familiar readers to gain background necessary to replicate the paper}{(yes/no)}
yes

\end{itemize}
\checksubsection{Theoretical Contributions}
\begin{itemize}

\question{Does this paper make theoretical contributions?}{(yes/no)}
no

	\ifyespoints{\vspace{1.2em}If yes, please address the following points:}
        \begin{itemize}
	
	\question{All assumptions and restrictions are stated clearly and formally}{(yes/partial/no)}
	Type your response here

	\question{All novel claims are stated formally (e.g., in theorem statements)}{(yes/partial/no)}
	Type your response here

	\question{Proofs of all novel claims are included}{(yes/partial/no)}
	Type your response here

	\question{Proof sketches or intuitions are given for complex and/or novel results}{(yes/partial/no)}
	Type your response here

	\question{Appropriate citations to theoretical tools used are given}{(yes/partial/no)}
	Type your response here

	\question{All theoretical claims are demonstrated empirically to hold}{(yes/partial/no/NA)}
	Type your response here

	\question{All experimental code used to eliminate or disprove claims is included}{(yes/no/NA)}
	Type your response here
	
	\end{itemize}
\end{itemize}

\checksubsection{Dataset Usage}
\begin{itemize}

\question{Does this paper rely on one or more datasets?}{(yes/no)}
yes

\ifyespoints{If yes, please address the following points:}
\begin{itemize}

	\question{A motivation is given for why the experiments are conducted on the selected datasets}{(yes/partial/no/NA)}
	yes

	\question{All novel datasets introduced in this paper are included in a data appendix}{(yes/partial/no/NA)}
	yes

	\question{All novel datasets introduced in this paper will be made publicly available upon publication of the paper with a license that allows free usage for research purposes}{(yes/partial/no/NA)}
	yes

	\question{All datasets drawn from the existing literature (potentially including authors' own previously published work) are accompanied by appropriate citations}{(yes/no/NA)}
	yes

	\question{All datasets drawn from the existing literature (potentially including authors' own previously published work) are publicly available}{(yes/partial/no/NA)}
	yes

	\question{All datasets that are not publicly available are described in detail, with explanation why publicly available alternatives are not scientifically satisficing}{(yes/partial/no/NA)}
	yes

\end{itemize}
\end{itemize}

\checksubsection{Computational Experiments}
\begin{itemize}

\question{Does this paper include computational experiments?}{(yes/no)}
yes

\ifyespoints{If yes, please address the following points:}
\begin{itemize}

	\question{This paper states the number and range of values tried per (hyper-) parameter during development of the paper, along with the criterion used for selecting the final parameter setting}{(yes/partial/no/NA)}
	yes

	\question{Any code required for pre-processing data is included in the appendix}{(yes/partial/no)}
	yes

	\question{All source code required for conducting and analyzing the experiments is included in a code appendix}{(yes/partial/no)}
	yes

	\question{All source code required for conducting and analyzing the experiments will be made publicly available upon publication of the paper with a license that allows free usage for research purposes}{(yes/partial/no)}
	yes
        
	\question{All source code implementing new methods have comments detailing the implementation, with references to the paper where each step comes from}{(yes/partial/no)}
	yes

	\question{If an algorithm depends on randomness, then the method used for setting seeds is described in a way sufficient to allow replication of results}{(yes/partial/no/NA)}
	yes

	\question{This paper specifies the computing infrastructure used for running experiments (hardware and software), including GPU/CPU models; amount of memory; operating system; names and versions of relevant software libraries and frameworks}{(yes/partial/no)}
	yes

	\question{This paper formally describes evaluation metrics used and explains the motivation for choosing these metrics}{(yes/partial/no)}
	yes

	\question{This paper states the number of algorithm runs used to compute each reported result}{(yes/no)}
	yes

	\question{Analysis of experiments goes beyond single-dimensional summaries of performance (e.g., average; median) to include measures of variation, confidence, or other distributional information}{(yes/no)}
	yes

	\question{The significance of any improvement or decrease in performance is judged using appropriate statistical tests (e.g., Wilcoxon signed-rank)}{(yes/partial/no)}
	yes

	\question{This paper lists all final (hyper-)parameters used for each model/algorithm in the paper’s experiments}{(yes/partial/no/NA)}
	yes

\end{itemize}
\end{itemize}
\ifreproStandalone
\end{document}
\fi
\end{document}